# Image De-Noising For Salt and Pepper Noise by Introducing New Enhanced Filter

Pranay Yadav, Vivek Kumar, Manju Jain, Atul Samadhiya, and Sandeep Jain

*Abstract*—When an image is formed, factors such as lighting (spectra, source, and intensity) and camera characteristics (sensor response, lenses) affect the appearance of the image. Therefore, the prime factor that reduces the quality of the image is noise. It hides the important details and information's of images. In order to enhance the qualities of image, the removal of noises become imperative and that should not at the cost of any loss of image information. Noise removal is one of the pre-processing stages of image processing. In this paper a new method for the enhancement of gray scale images is introduced, when images are corrupted by fixed valued impulse noise (salt and pepper noise). The proposed methodology ensures a better output for low and medium density of fixed value impulse noise as compare to the other famous filters like Standard Median Filter (SMF), Decision Based Median Filter (DBMF) and Modified Decision Based Median Filter (MDBMF) etc. The main objective of the proposed method was to improve peak signal to noise ratio (PSNR), visual perception and reduction in blurring of image. The proposed algorithm replaced the noisy pixel by trimmed mean value. When previous pixel values, 0's and 255's are present in the particular window and all the pixel values are 0's and 255's then the remaining noisy pixels are replaced by mean value. The gray-scale image of mandrill and Lena were tested via proposed method. The experimental result shows better peak signal to noise ratio (PSNR), mean square error values with better visual and human perception.

*Keywords*— Blurring, Human and visual perception, Modified Nonlinear filter, Salt and Pepper noise, Mean Square Error, Peak Signal to Noise Ratio.

## I. Introduction

IN the field of image processing, digital images very often get corrupted by several kinds of noise during the process of image acquisition. Primarily, because of the reasons like malfunctioning of pixels in camera sensors, faulty memory locations in hardware, or transmission in a noisy channel [1]. In addition, these are also the main reasons responsible for generation of the impulse noise in digital world. In the field of image processing, digital images are mainly corrupted by the impulse noise [2]. There are two types of impulse noise, namely, the salt-and-pepper noise also known as the fixed valued impulse noise and the random-valued impulse noise [3]. Impulse noise is one the most severe noise which usually affects the images. Researchers are involved in the field of image de-noising in order to find out effective method, capable of preserving the image details, reducing the noise of digital images and ensuring the quality of the image. Image quality measurement is analyzed by image parameters like peak to single noise ratio (PSNR), mean square error (MSE), image enhancement factor (IEF), but in case of image processing one more thing of utmost importance is human perception [4]. In this paper focus is kept upon salt and pepper noise. The salt and pepper noise corrupted pixels of image take either maximum or minimum pixel value Salt and pepper noise. Fixed valued impulse noise is producing two gray level values 0 and 255. Random valued impulse noise will produce impulses whose gray level value lies within a predetermined range. The random value impulse noise is between 0 and 255.

Generally the spatial domain filters have a detection stage which identifies the noisy and noise free pixels of the corrupted image, after that noise removal part removes the noise from the corrupted image under process while preserving the other important detail of image [5].

Initially standard median filter was popularly used, but later on switching based median filters came into existence which provides better results. Any other result oriented standard median filters are, weighted median filter, SD-ROM filter, centre weighted median filter, adaptive median filter, rank order median filter and many other improved filters. The performance of median filters also depends on the size of window of the filter. Larger window has the great noise suppression capability, but image details (edges, corners, fine lines) preservation is limited, while a smaller window preserves the details, but it will cause the reduction in noise suppression. Noise detection is a vital part of a filter, so it is necessary to detect whether the pixel is noise or noise free. However, further reduction in computational complexity is enviable.

## II. Proposed Methodology

The proposed method deploys the enhancement by Modified Non-linear Filter (MNF) [03] algorithm. In this method first task is to detect the noisy pixels in the corrupted image. For detection of noisy pixels verifying the condition whether targeted pixel lies. If pixels are between maximum [255] and minimum [0] gray level values, then it is a noise

Pranay Yadav is with TIT College, RGTU, Bhopal-462021, INDIA (E-mail: pranaymedc@gmail.com).
Vivek Kumar is Lecturer and TPO with Laxmipati Group of Institutions, RGTU, Bhopal-462021, INDIA (Phone: +91 9098374992; e-mail: kunwarv4@gmail.com).
Manju Jain was with RKDF College, RGTU, Bhopal-462021, INDIA (E-mail: manjujain87@gmail.com).
Atul Samadhiya is with IES college Bhopal-462021, INDIA (e-mail: aswoodstock40@gmail.com).
Sandeep Jain is with LIST college, RGTU, Bhopal-462021, INDIA (e-mail: jainsandeep10@hotmail.com).

free pixel, else pixel is said to be corrupted or noisy. Now we have processed only with the corrupted pixels to restore with noise free pixels. Further un-corrupted pixels are left unaffected. In the next steps we use Proposed Improved Mean filter (RMF) is elucidated as follows.

ALGORITHM

Step 1: Initially in the very first step an image was taken and fixed valued impulses noise (Salt and Pepper noise) was applied on the image.

Step 2: In the second it had been checked whether the pixels falls in between 0 to 255 ranges or not, consequently two cases were arises

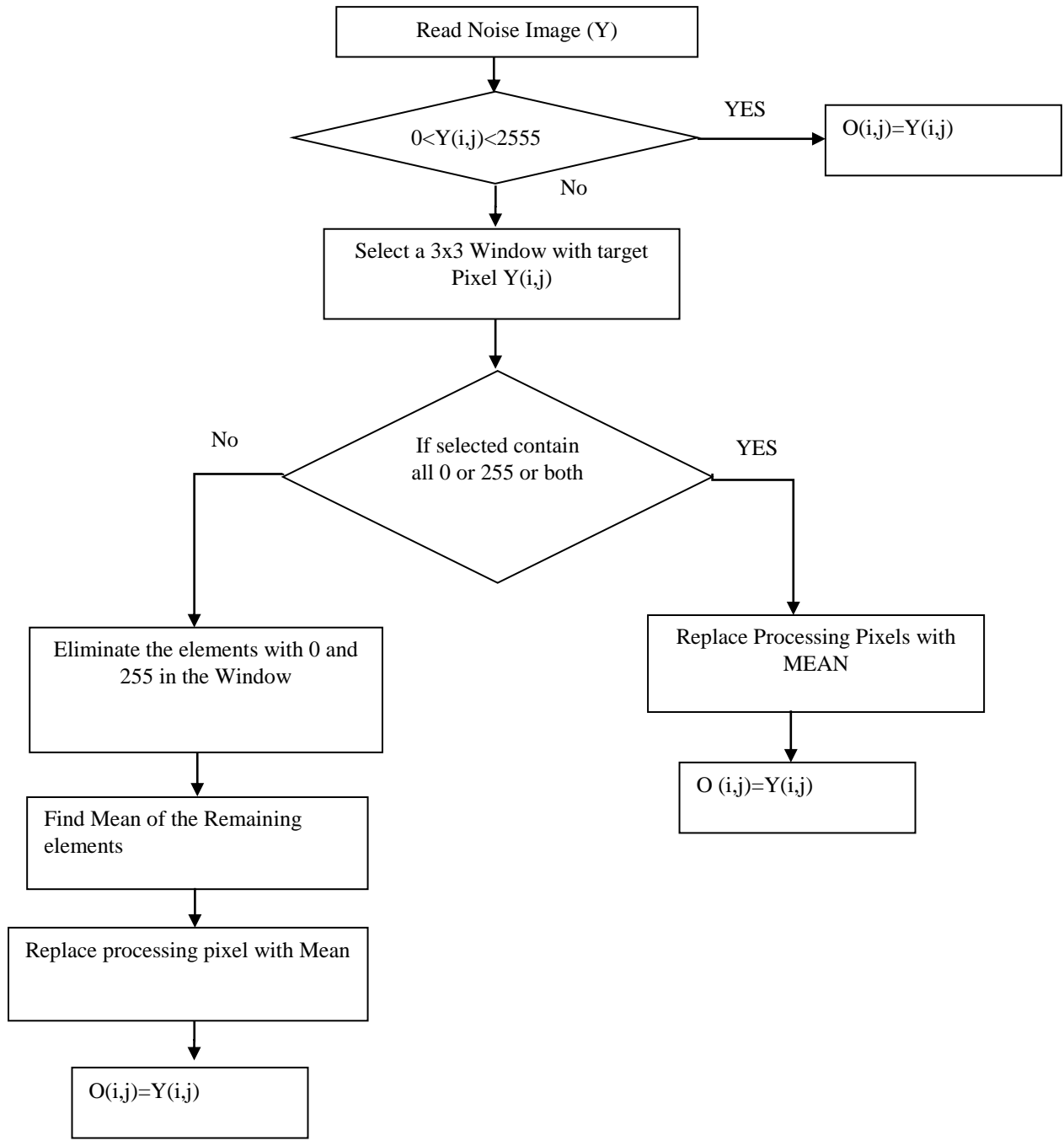

Fig. 1 Flow-Chart diagram of Proposed Method

X (i,j) = 0<Y (i0 the image size and Y (i,,j) all image targeted pixels.

Case 1- If pixels are between 0< Y (i,j)<255 then, they are noise free and move to restoration image.

Case 2- If the pixels are not lying between in the range then they it is moved towards step 3.

Step 3: In the third step work is done on noisy pixel of step-2. A window of size 3 x 3 of image was selected. Assuming that the targeted noisy pixels were W (ij). It was processed in the next step.

Step 4: If the preferred window does not contains all elements as 0's and 255's then remove all the 0's and 255's from the window, and send to restoration image. Now find the mean of the remaining pixels was found. W (i,j) was replaced with the mean value. This removed noise of image was restored as de-noised image at the last step.

Y (ij) = [00] condition true send to O (i,j) for Restoration

Y(ij) = [255] condition true send to O(i,j) for Restoration

[Cal. Mean remain (Y (ij)) pixels] = replace by O (i,j)

Step 5: Steps one to three were repeated until all pixels in the whole image was processed. Hence a better de-noised image was obtained with improved PSNR, MSE and also shows a better image with very low blurring and improved visual and human perception.

The proposed algorithm's block based flowchart is shown in figure. In the flowchart all the conditions have been mentioned with utmost clarity.

### III. RESULTS

The result of the proposed method for removal of fixed valued impulse noise is shown in this section. For simulation of proposed method we have to use MATLAB 8.0 software. To perform our new approach we have to take a 'Leena' image size 256X256 as a reference image for testing purpose. The testing images are artificially corrupted by Salt and Pepper impulse noise by using MATLAB and images are corrupted by different noise density varying from 10 to 90 %. The performance of the proposed algorithm is tested for gray scale image.

De-noising performances are quantitatively measured by the PSNR and MSE as defined in (1) and (3), respectively:

The PSNR is expressed as:

$$PSNR = 10\log_{10}\frac{(255)^2}{MSE} \quad (1)$$

Where MSE (Mean Square Error) is

$$MSE = \frac{\sum_{i=1}^{m}\sum_{j=1}^{n}\{Y(i,j)-\hat{Y}(i,j)\}^2}{m \times n} \quad (2)$$

Where MSE is acronym of mean square error.

The PSNR and MSE values of the proposed algorithm are compared with other existing algorithms by variable noise density of 10% to 50%. Table 1 shows the comparison of PSNR values of different de-noising methods for Lena image.

TABLE I
VALUES OF PSNR FOR VARYING NOISE DENSITY

| De-noising Method | PSNR in dB with varying percentage of noise | | | | |
|---|---|---|---|---|---|
| | Noise density | | | | |
| | 10% | 20% | 30% | 40% | 50% |
| MF | 28.4938 | 25.7542 | 21.8465 | 18.4076 | 14.734 |
| AMF | 21.9845 | 21.9297 | 21.4735 | 21.4735 | 20.6542 |
| PSMF | 30.6494 | 28.2089 | 25.559 | 22.6909 | 19.4425 |
| DBA | 36.7565 | 33.2606 | 30.5659 | 28.2609 | 26.2846 |
| MDBA | 36.7569 | 33.2607 | 30.5308 | 28.2981 | 26.2503 |
| MDBUTMF | 38.129 | 34.6005 | 32.1427 | 32.0886 | 28.2175 |
| New Approach | 34.2762 | 34.2762 | 34.2762 | 34.2762 | 34.2762 |

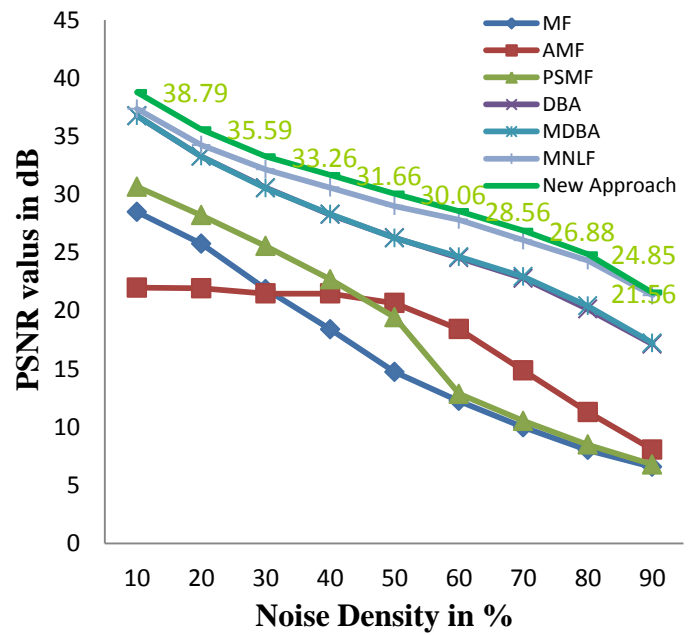

Fig.2 PSNR (dB) Vs Noise Density %

TABLE II
VALUES OF PSNR, MSE AND TIME ELAPSED FOR VARYING PERCENTAGE OF NOISE OF PROPOSED MEAN FILTER.

| Noise percentage (%), | PSNR | MSE | Elapsed time (in seconds) |
|---|---|---|---|
| 10 | 34.2762 | 24.2920 | 3.0 |
| 20 | 31.1540 | 49.8513 | 3.1 |
| 30 | 29.2863 | 76.6391 | 3.69 |
| 40 | 27.7298 | 109.6742 | 4.66 |
| 50 | 26.5649 | 143.4142 | 6.50 |
| 60 | 25.5204 | 182.4069 | 6.89 |
| 70 | 24.3077 | 241.1607 | 7.60 |
| 80 | 23.0167 | 324.6479 | 8.33 |
| 90 | 21.4067 | 470.3241 | 9.09 |

The proposed new approach shows a better result as compare to other existing algorithms at different noise densities as shown in Table-1. Our method not only shows a better result in terms of PSNR and MSE, but also show a good result in visual and human perception is also shown in the these shows the visual quality of the image.

Graphical plots of PSNR values of different noise density compression with different filters against noise densities for Lena image is shown in Figure 3.

The results in the Table 4 clearly show that the PSNR of the proposed method is much improved at different density of noise.

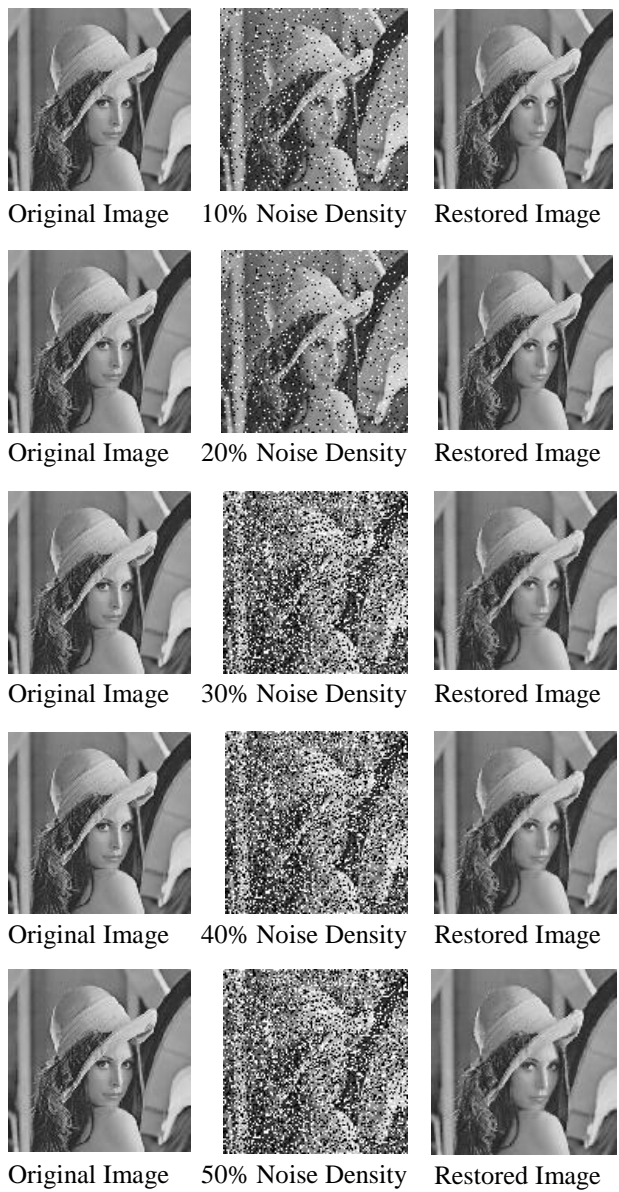

Original Image    10% Noise Density    Restored Image

Original Image    20% Noise Density    Restored Image

Original Image    30% Noise Density    Restored Image

Original Image    40% Noise Density    Restored Image

Original Image    50% Noise Density    Restored Image

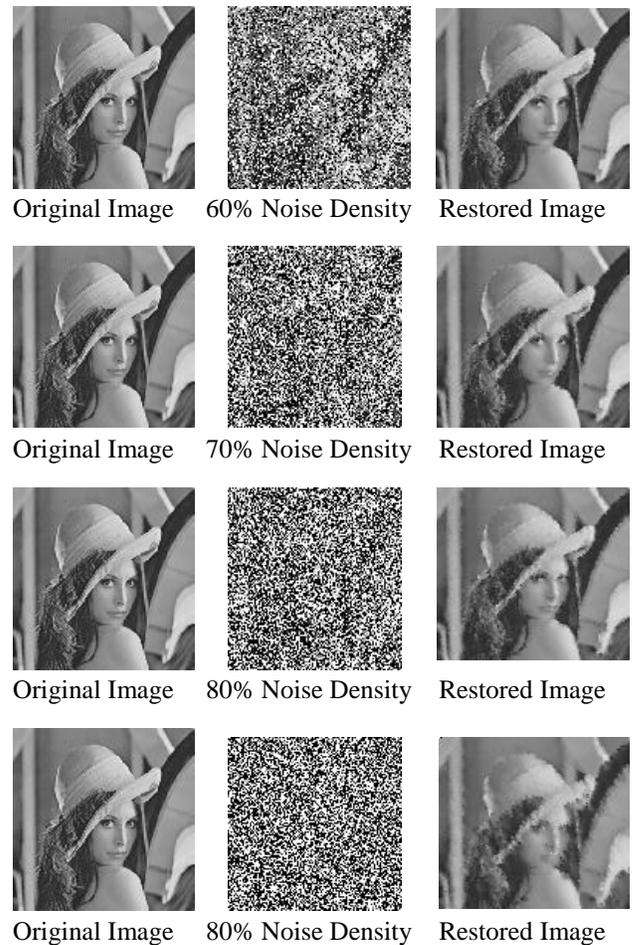

Original Image    60% Noise Density    Restored Image

Original Image    70% Noise Density    Restored Image

Original Image    80% Noise Density    Restored Image

Original Image    80% Noise Density    Restored Image

Fig. 3 Experimental results of proposed method for Lena image with noise density being varied from (10-90) %.

## IV. RESULTS

A new algorithm has been proposed to deal with the problems, namely, poor image enhancement at high noise density, which is frequently enhanced in the PSNR. In this paper Improved Mean Filtering is used for enhancing the peak signal to noise ratio (PSNR) and MSE both. The performances of proposed 'Robust Mean Filter' (RMF) are quantitatively vies as well as the visual and human perception vies shows better result in both conditions as compared to other existing filters. Results reveal that the proposed filter exhibits better performance in comparison with MF, AMF, DBA, MDBA, MDBUTMF, MNF filters in terms of higher PSNR and MSE. Indifference to AMF and other existing algorithms, the new algorithm uses a small 3x3 window having only eight neighbors of the corrupted pixel that have higher connection; this provides more edge information, more important to better edge preservation as well as more better human and visual perception. The new algorithm filter also shows reliable and stable performance across a different range of noise densities varying from 10% - 90%.

The performance of the proposed method has been tested at low, medium and high noise densities on gray scales. In fact at high noise density levels the new proposed algorithm gives better performance as compare with other existing de-noising filters.

REFERENCES

[1] Raymond H. Chan, Chung-Wa Ho, and Mila Nikolova " Salt-and-Pepper Noise Removal by Median-Type Noise Detectors and Detail-Preserving Regularization" IEEE TRANSACTIONS ON IMAGE PROCESSING, VOL. 14, NO. 10, OCTOBER 2005 1479.

[2] S.Esakkirajan,T.Veerakumar, Adabala N. Subramanyam and C. H. PremChand. "Removal of High Density Salt and Pepper Noise Through Modified Decision Based Unsymmetric Trimmed Median Filter." *IEEE Signal Processing Letters,* VOL. 18, NO. 5, MAY 2011.

[3] P. E. Ng and K. K. Ma, "A switching median filter with boundary discriminative noise detection for extremely corrupted images,"*IEEE Trans. Image Process.*, vol. 15, no. 6, pp. 1506–1516, Jun. 2006. http://dx.doi.org/10.1109/TIP.2005.871129

[4] T.Sunilkumar, A.Srinivas, M.Eswae Reddy and Dr. G.Ramachandren Reddy "Removal of high density impulse noise through modified non-linear filter" Journal of Theoretical and Applied Information Technology.. Vol. 47 No.2,20th January 2013 ISSN: **1992-8645,** E-ISSN: **1817-3195.**

[5] S. Zhang and M. A. Karim, "A new impulse detector for switching median filters," *IEEE Signal Process. Lett.*, vol. 9, no. 11, pp. 360–363, Nov. 2002. http://dx.doi.org/10.1109/LSP.2002.805310

[6] H. Hwang and R. A. Hadded, "Adaptive median filter: New algorithms and results," *IEEE Trans. Image Process.*, vol. 4, no. 4, pp. 499–502, Apr. 1995. http://dx.doi.org/10.1109/83.370679

[7] K. S. Srinivasan and D. Ebenezer, "A new fast and efficient decision based algorithm for removal [7] V. Jayaraj and D. Ebenezer, "A new switching-based median filtering scheme and algorithmfor removal of high-density salt and pepper noise in image," *EURASIP J. Adv. Signal Process.*, 2010

[8] T.A. Nodes and N.C. Gallagher, Jr., "The output distribution of median type filters," IEEE Trans. Communication., 32(5): 532-541, 1984. http://dx.doi.org/10.1109/TCOM.1984.1096099

[9] V. Jayaraj and D. Ebenezer, "A new switching-based median filtering scheme and algorithm for removal of high density salt and pepper noise in removal of high-density salt and pepper noise in image," *EURASIP J. Adv. Signal Process.*, 2010. http://dx.doi.org/10.1155/2010/690218

[10] K. Aiswarya, V. Jayaraj, and D. Ebenezer, "A new and efficient algorithm for the removal of high density salt and pepper noise in images and videos," in *Second Int. Conf. Computer Modeling and Simulation*, 2010, pp. 409–413.

[11] V. Jayaraj and D. Ebenezer, "A new switching-based median filtering scheme and algorithmfor removal of high-density salt and pepper noise in image," *EURASIP J. Adv. Signal Process.* 2010. http://dx.doi.org/10.1155/2010/690218

[12] J. Astola and P. Kuosmaneen*, Fundamentals of Nonlinear Digital Filtering*. Boca Raton, FL: CRC, 1997